\documentclass[runningheads]{llncs}
\usepackage{graphicx}
\usepackage{comment}
\usepackage{amsmath,amssymb} 
\usepackage{color}
\usepackage{caption}
\usepackage[colorlinks=true]{hyperref}

\usepackage{amsfonts}
\usepackage{array}
\usepackage{booktabs}
\usepackage{color}
\usepackage{enumitem}
\usepackage{float}
\usepackage{mathtools}
\usepackage{multirow}
\usepackage{setspace}

\newcolumntype{L}[1]{>{\raggedright\let\newline\\\arraybackslash\hspace{0pt}}m{#1}}
\newcolumntype{C}[1]{>{\centering\let\newline\\\arraybackslash\hspace{0pt}}m{#1}}
\newcolumntype{R}[1]{>{\raggedleft\let\newline\\\arraybackslash\hspace{0pt}}m{#1}}
\newcommand\Tstrut{\rule{0pt}{2.6ex}}         
\newcommand\Bstrut{\rule[-1.3ex]{0pt}{0pt}}   

\setlist[itemize]{noitemsep, topsep=0pt}
\setlist[enumerate]{noitemsep, topsep=0pt}

\newcommand{\smallfont}{\fontsize{8}{9.6}\selectfont}

\newcommand{\parens}[1]{\left(#1\right)}
\newcommand{\braces}[1]{\left\{#1\right\}}
\newcommand{\bracks}[1]{\left[#1\right]}

\newcommand{\norm}[1]{\left\Vert#1\right\Vert}

\newcommand{\samplei}{^{\parens{i}}}
\newcommand{\raisedth}[1]{#1^{\textrm{th}}}

\newcommand\blfootnote[1]{%
  \begingroup
  \renewcommand\thefootnote{}\footnote{#1}%
  \addtocounter{footnote}{-1}%
  \endgroup
}

\begin{document}
\pagestyle{headings}
\mainmatter

\title{Take an Emotion Walk: Perceiving Emotions from Gaits Using Hierarchical Attention Pooling and Affective Mapping} 


\titlerunning{Take an Emotion Walk}
%
\author{Uttaran Bhattacharya\orcidID{0000-0003-2141-9276}$^{,}$\inst{1} \and
Christian Roncal\orcidID{0000-0003-2772-5071}$^{,}$\inst{1} \and
Trisha Mittal\orcidID{0000-0003-3558-6518}$^{,}$\inst{1} \and
Rohan Chandra\orcidID{0000-0003-4843-6375}$^{,}$\inst{1} \and
Kyra Kapsaskis\orcidID{0000-0003-2063-6156}$^{,}$\inst{2} \and
Kurt Gray\orcidID{0000-0001-5816-2676}$^{,}$\inst{2} \and
Aniket Bera\orcidID{0000-0002-0182-6985}$^{,}$\inst{1} \and
Dinesh Manocha\orcidID{0000-0001-7047-9801}$^{,}$\inst{1}
}
\authorrunning{U. Bhattacharya et al.}
%
\institute{University of Maryland, College Park, MD 20742, USA \and
University of North Carolina, Chapel Hill, NC 27599, USA
}
\maketitle

\blfootnote{This project has been supported by ARO grant W911NF-19-1-0069.}
\blfootnote{Code and additional materials in project webpage: \url{https://gamma.umd.edu/taew}}
\setcounter{footnote}{0}

\begin{abstract}
    We present an autoencoder-based semi-supervised approach to classify perceived human emotions from walking styles obtained from videos or motion-captured data and represented as sequences of 3D poses. Given the motion on each joint in the pose at each time step extracted from 3D pose sequences, we hierarchically pool these joint motions in a bottom-up manner in the encoder, following the kinematic chains in the human body. We also constrain the latent embeddings of the encoder to contain the space of psychologically-motivated affective features underlying the gaits. We train the decoder to reconstruct the motions per joint per time step in a top-down manner from the latent embeddings. For the annotated data, we also train a classifier to map the latent embeddings to emotion labels. Our semi-supervised approach achieves a mean average precision of 0.84 on the Emotion-Gait benchmark dataset, which contains both labeled and unlabeled gaits collected from multiple sources. We outperform current state-of-art algorithms for both emotion recognition and action recognition from 3D gaits by 7\%--23\% on the absolute. More importantly, we improve the average precision by 10\%--50\% on the absolute on classes that each makes up less than 25\% of the labeled part of the Emotion-Gait benchmark dataset.
\end{abstract}

\section{Introduction}\label{sec:intro}
Humans perceive others' emotions through verbal cues such as speech~\cite{prosodic1,prosodic2}, text~\cite{text1,text2}, and non-verbal cues such as eye-movements~\cite{eye2}, facial expressions~\cite{AU}, tone of voice, postures~\cite{posture1}, walking styles~\cite{gait_psych4}, etc.
Perceiving others' emotions shapes people's interactions and experiences when performing tasks in collaborative or competitive environments~\cite{secret_life_of_brain}.
Given this importance of perceived emotions in everyday lives, there has been a steady interest in developing automated techniques for perceiving emotions from various cues, with applications in affective computing, therapy, and rehabilitation~\cite{rehab}, robotics~\cite{robotics,proxemo}, surveillance~\cite{surveillance,liarswalk}, audience understanding~\cite{audience_understanding}, and character generation~\cite{char_gen}.

While there are multiple non-verbal modalities for perceiving emotions, in our work, we only observe people's styles of walking or their gaits, extracted from videos or motion-captured data. Perceived emotion recognition using any non-verbal cues is considered to be a challenging problem in both psychology and AI, primarily because of the unreliability in the cues, arising from sources such as ``mock'' expressions~\cite{unreliability1}, expressions affected by the subject's knowledge of an observer~\cite{unreliability2}, or even self-reported emotions in certain scenarios~\cite{unreliability3}. However, gaits generally require less conscious initiation from the subjects and therefore tend to be more reliable cues. Moreover, studies in psychology have shown that observers were able to perceive the emotions of walking subjects by observing features such as arm swinging, stride lengths, collapsed upper body, etc.~\cite{gait_psych3,gait_psych4}.

Gaits have been widely used in computer vision for many applications, including action recognition~\cite{stgcn,dgnn,msg3d} and perceiving emotions~\cite{tanmay_emotions,eva,step,emoticon}. However, there are a few key challenges in terms of designing machine learning methods for emotion recognition using gaits: 
\begin{itemize}[label=\textbullet]
    \item Methods based on hand-crafted biomechanical features extracted from human gaits often suffer from low prediction accuracy~\cite{crenn2016body,venture2014recognizing}.
    \item Fully deep-learned methods~\cite{tanmay_emotions,step} rely heavily on sufficiently large sets of annotated data. Annotations are expensive and tedious to collect due to the variations in scales and motion trajectories~\cite{discrimnet}, as well as the inherent subjectivity in perceiving emotions~\cite{step}. The benchmark dataset for emotion recognition, Emotion-Gait~\cite{step}, has around $4,000$ data points of which more than $53\%$ are unlabeled.
    \item Conditional generative methods are useful for data augmentation, but current methods can only generate data for short time periods~\cite{orange_duck,video_ac_gen_2} or with relatively low diversity~\cite{quaternet,gait_ac_gen_1,video_ac_gen_1,step}.
\end{itemize}
On the other hand, acquiring poses from videos and MoCap data is cheap and efficient, leading to the availability of large-scale pose-based datasets~\cite{cmu_mocap,human3.6m,kinetics,ntu_rgbd}.
Given the availability of these unlabeled gait datasets and the sparsity of gaits labeled with perceived emotions, there is a need to develop automatic methods that can utilize these datasets for emotion recognition.

\noindent\textbf{Main Contributions:}
We present a semi-supervised network that accepts 3D pose sequences of human gaits extracted from videos or motion-captured data and predicts discrete perceived emotions, such as happy, angry, sad, and neutral. Our network consists of an unsupervised autoencoder coupled with a supervised classifier. The encoder in the unsupervised autoencoder hierarchically pools attentions on parts of the body. It learns separate intermediate feature representations for the motions on each of the human body parts (arms, legs, and torso) and then pools these features in a bottom-up manner to map them to the latent embeddings of the autoencoder. The decoder takes in these embeddings and reconstructs the motion on each joint of the body in a top-down manner.

We also perform affective mapping: we constrain the space of network-learned features to subsume the space of biomechanical affective features~\cite{aff_features} expressed from the input gaits. These affective features contain useful information for distinguishing between different perceived emotions.
Lastly, for the labeled data, our supervised classifier learns to map the encoder embeddings to the discrete emotion labels to complete the training process. To summarize, we contribute:
\begin{itemize}[label=\textbullet]
    \item \textbf{A semi-supervised network}, consisting of an autoencoder and a classifier, that are trained together to predict discrete perceived emotions from 3D pose sequences of gaits of humans.
    \item \textbf{A hierarchical attention pooling module} on the autoencoder to learn useful embeddings for unlabeled gaits, which improves the mean average precision (mAP) in classification by 1--17\% on the absolute compared to state-of-the-art methods in both emotion recognition and action recognition from 3D gaits on the Emotion-Gait benchmark dataset.
    \item \textbf{Subsuming the affective features} expressed from the input gaits in the space of learned embeddings. This improves the mAP in classification by 7--23\% on the absolute compared to state-of-the-art methods.
\end{itemize}
We observe the performance of our network improves linearly as more unlabeled data is used for training. More importantly, we report a 10--50\% improvement on average precision on the absolute for emotion classes that have fewer than 25\% labeled samples in the Emotion-Gait dataset~\cite{step}.

\section{Related Work}\label{sec:rw}
We briefly review prior work in classifying perceived emotions from gaits, as well as the related task of action recognition and generation from gaits.

\noindent\textbf{Detecting Perceived Emotions from Gaits.} Experiments in psychology have shown that observers were able to identify sadness, anger, happiness, and pride by observing gait features such as arm swinging, long strides, erect posture, collapsed upper body, etc.~\cite{gait_psych1,gait_psych2,gait_psych3,gait_psych4}. This, in turn, has led to considerable interest from both the computer vision and the affective computing communities in detecting perceived emotions from recorded gaits. Early works exploited different gait-based affective features to automatically detect perceived emotions~\cite{karg2010recognition,venture2014recognizing,crenn2016body,daoudi2017emotion}. More recent works combined these affective features with features learned from recurrent~\cite{tanmay_emotions} or convolutional networks~\cite{step} to significantly improve classification accuracies.

\noindent\textbf{Action Recognition and Generation.} There are large bodies of recent work on both gait-based supervised action recognition~\cite{video_ac_reg_1,video_ac_reg_2,stgcn,recurrent_action,gc_lstm,dgnn,2sagcn,msg3d}, and gait-based unsupervised action generation~\cite{video_ac_gen_1,gait_ac_gen_1,orange_duck,quaternet}. These methods make use of RNNs or CNNs, including GCNs, or a combination of both, to achieve high classification accuracies on benchmark datasets such as Human3.6M~\cite{human3.6m}, Kinetics~\cite{kinetics}, NTU RGB-D~\cite{ntu_rgbd}, and more. On top of the deep-learned networks, some methods have also leveraged the kinematic dependencies between joints and bones~\cite{dgnn}, dynamic movement-based features~\cite{2sagcn}, and long-range temporal dependencies~\cite{msg3d}, to further improve performance. A comprehensive review of recent methods in kinect-based action recognition is available in~\cite{kinect_review}.

RNN and CNN-based approaches have been extended to semi-supervised classification as well~\cite{recurrent_semisup,conv_semisup,lhd,phd}. These methods have also added constraints on limb proportions, movement constraints, and exploited the autoregressive nature of gait prediction to improve their generative and classification components.

Generative methods have also exploited full sequences of poses to directly generate full test sequences~\cite{pose_guided_gait1,pose_guided_gait2}. Other approaches have used constraints on limb movements~\cite{discrimnet}, action-specific trajectories~\cite{orange_duck}, and the structure and kinematics of body joints~\cite{quaternet}, to improve the naturalness of generated gaits.

In our work, we learn latent embeddings from gaits by exploiting the kinematic chains in the human body~\cite{human_kinematics} in a hierarchical fashion. Inspired by prior works in emotion perception from gaits, we also constrain our embeddings to contain the space of affective features expressed from gaits, to improve our average precision, especially on the rarer classes.

\section{Approach}\label{sec:approach}
Given both labeled and unlabeled 3D pose sequences for gaits, our goal is classify all the gaits into one or more discrete perceived emotion labels, such as happy, sad, angry, etc. We use a semi-supervised approach to achieve this, by combining an autoencoder with a classifier, as shown in Fig.~\ref{fig:network}. We denote the set of trainable parameters in the encoder, decoder, and classifier with $\theta$, $\psi$, and $\phi$ respectively. We first extract the rotation per joint from the first time step to the current time step in the input sequences (details in Sec.~\ref{subsec:preprocessing}). We then pass these rotations through the encoder, denoted with $f_\theta\parens{\cdot}$, to transform the input rotations into features in the latent embedding space. We pass these latent features through the decoder, denoted with $f_\psi\parens{\cdot}$, to generate reconstructions of the input rotations. If training labels are available, we also pass the encoded features through the fully-connected classifier network, denoted with $f_\phi\parens{\cdot}$, to predict the probabilities of the labels. We define our overall loss function as
\begin{equation}
        \mathcal{C}\parens{\theta, \phi, \psi} = \sum_{i=1}^M I_y\samplei\mathcal{C}_{CL}\parens{y\samplei, f_{\phi\circ\theta}\parens{D\samplei}} + \mathcal{C}_{AE}\parens{D\samplei, f_{\psi\circ\theta}\parens{D\samplei}},
    \label{eq:semisup_loss}
\end{equation}
where $f_{b\circ a}\parens{\cdot} := f_b\parens{f_a\parens{\cdot}}$ denotes the composition of functions, $I_y\samplei$ is an indicator variable denoting whether the $\raisedth{i}$ data point has an associated label $y\samplei$, $M$ is the number of gait samples, $\mathcal{C}_{CL}$ denotes the classifier loss detailed in Sec.~\ref{subsec:cf_loss}, and $\mathcal{C}_{AE}$ denotes the autoencoder loss detailed in Sec.~\ref{subsec:ae_loss}. For brevity of notation, we will henceforth use $\hat{y}\samplei := f_{\phi\circ\theta}\parens{D\samplei}$ and $\hat{D}\samplei := f_{\psi\circ\theta}\parens{D\samplei}$.

\begin{table}[t]
    \begin{minipage}{0.38\linewidth}
        \centering
        \includegraphics[width=\columnwidth]{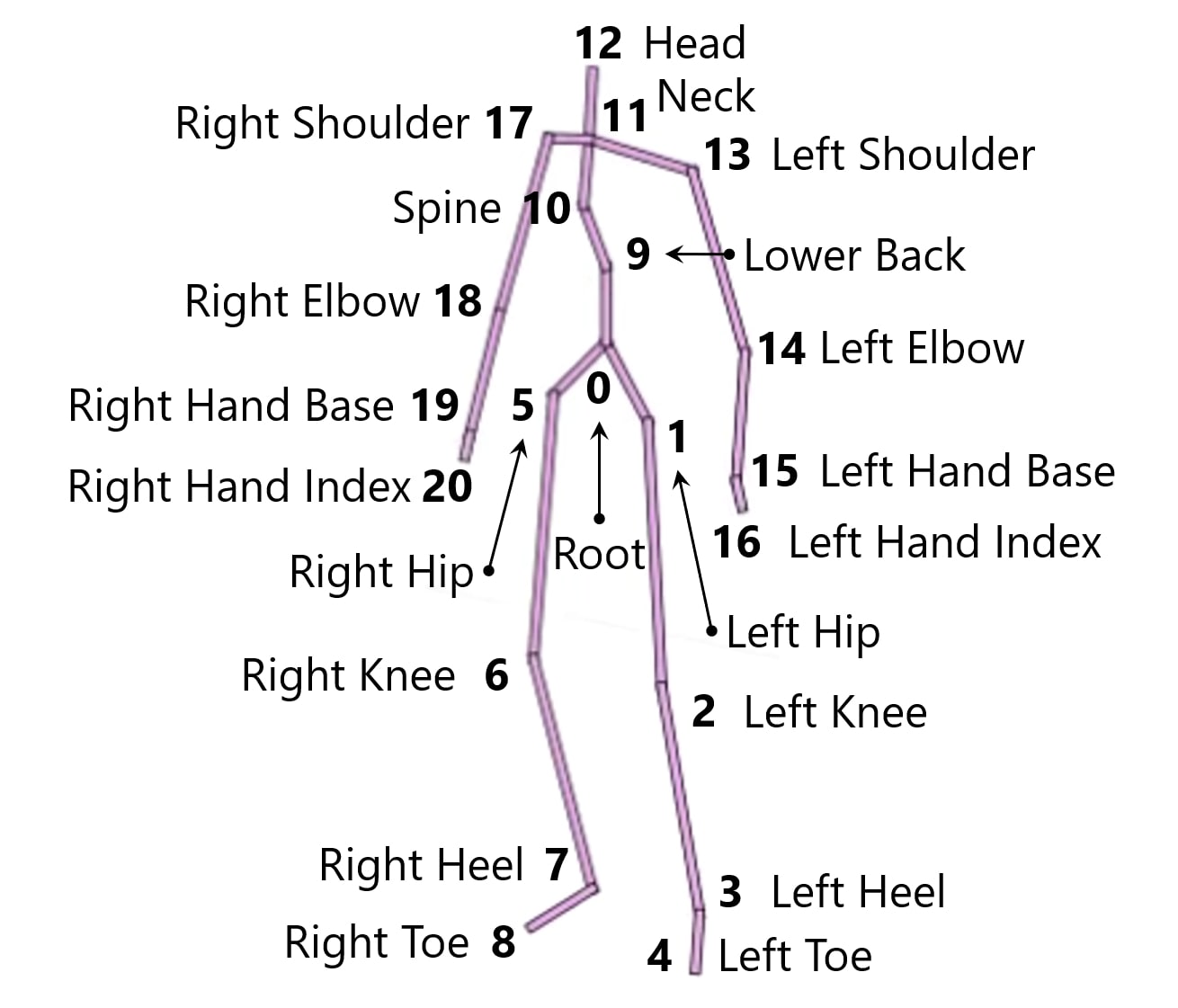}
        \captionof{figure}{\smallfont\textbf{3D pose model.} The names and numbering of the $21$ joints in the pose follow the nomenclature in the ELMD dataset~\cite{elmd}.}
        \label{fig:pose}
    \end{minipage}
    \hfill
    \begin{minipage}{0.60\linewidth}
        \caption{\smallfont\textbf{Affective Features.} List of the $18$ pose affective features that we use to describe the affective feature space for our network.}
        \label{tab:aff_features}
        \centering\tiny
        \resizebox{0.8\columnwidth}{!}{%
            \begin{tabular}{L{0.9cm}L{5.4cm}}
                \toprule
                \multirow{11}{0.9cm}{Angles between} & shoulders at lower back\\
                & hands at root\\
                
                & left shoulder and hand at elbow \\
    
                & right shoulder and hand at elbow \\
    
                & head and left shoulder at neck \\
    
                & head and right shoulder at neck \\
    
                & head and left knee at root \\
    
                & head and right knee at root \\
    
                & left toe and right toe at root \\
    
                & left hip and toe at knee \\
    
                & right hip and toe at knee \\
                \midrule
                \multirow{4}{0.9cm}{Distance ratios between} & left hand index (LHI) to neck and LHI to root \\
    
                & right-hand index (RHI) to neck and RHI to root \\
    
                & LHI to RHI and neck to root \\
    
                & left toe to right toe and neck to root \\
                \midrule
                Area($\Delta$) & $\Delta$ shoulders to lower back and $\Delta$ shoulders to root \\
    
                ratios & $\Delta$ hands to lower back and $\Delta$ hands to root \\
    
                between & $\Delta$ hand indices to neck and $\Delta$ toes to root \\
                \bottomrule
            \end{tabular}
        }
    \end{minipage}
\end{table}

\subsection{Representing Emotions}\label{subsec:represent_emotions}
The Valence-Arousal-Dominance (VAD) model~\cite{vad} is used for representing emotions in a continuous space. This model assumes three independent axes for valence, arousal, and dominance values, which collectively indicate an observed emotion. Valence indicates how pleasant (vs. unpleasant) the emotion is, arousal indicates how much the emotion is tied to high (vs. low) physiological intensity, and dominance indicates how much the emotion is tied to the assertion of high (vs. low) social status. For example, discrete emotion terms such as happy indicate high valence, medium arousal, and low dominance, angry indicate low valence, high arousal, and high dominance, and sad indicate low valence, low arousal, and low dominance.

On the other hand, these discrete emotion terms are easily understood by non-expert annotators and end-users. As a result, most existing datasets for supervised emotion classification consist of discrete emotion labels, and most supervised methods report performance on predicting these discrete emotions. In fact, discrete emotions can actually be mapped back to the VAD space through various known transformations~\cite{discrete_to_continuous_1,discrete_to_continuous_2}. Given these factors, we choose to use discrete emotion labels in our work as well. We also note that human observers have been reported to be most consistent in perceiving emotions varying primarily on the arousal axis, such as happy, sad, and angry~\cite{critical_gait_features,effort_shape}. Hence we work with the four emotions, happy, sad, angry, and neutral.

\subsection{Representing the Data}\label{subsec:preprocessing}
Given the 3D pose sequences for gaits, we first obtain the rotations per joint per time step. We denote a gait as $G = \braces{\parens{x_j^t, y_j^t, z_j^t}}_{j=1, t=1}^{J, T}$, consisting of the 3D positions of $J$ joints across $T$ time steps. We denote the rotation of joint $j$ from the first time step to time step $t$ as $R_j^t \in \mathbb{SO}\parens{3}$. We represent these rotations as unit quaternions $q_j^t \in \mathbb{H} \subset \mathbb{R}^4$, where $\mathbb{H}$ denotes the space of unit 4D quaternions. As stated in~\cite{quaternet}, quaternions are free of the gimbal-lock problem, unlike other common representations such as Euler angles or exponential maps~\cite{exponential_maps}. We enforce the additional unit norm constraints for these quaternions when training our autoencoder. We represent the overall input to our network as $D\samplei := \braces{q_j^t}_{j=1, t=1}^{J, T} \in \mathbb{H}^{J\times T}$.

\subsection{Using Perceived Emotions and Constructing Classifier Loss}\label{subsec:cf_loss}
Observers' perception of emotions in others depends heavily influenced by their own personal, social, and cultural experiences, making emotion perception an inherently subjective task~\cite{critical_gait_features,gait_psych4}. Consequently, we need to keep track of the differences in the perceptions of different observers. We do this by assigning multi-hot emotion labels to each input gait.

We assume that the given labeled gait dataset consists of $C$ discrete emotion classes. The raw label vector $L\samplei$ for the $\raisedth{i}$ gait is a probability vector where the $\raisedth{l}$ element denotes the probability that the corresponding gait is perceived to have the $\raisedth{l}$ emotion. Specifically, we assume $L\samplei \in \bracks{0, 1}^C$ to be given as $L\samplei = \begin{bmatrix} p_1 & p_2 & \dots p_C \end{bmatrix}^\top$, where $p_l$ denotes the probability of the $\raisedth{l}$ emotion and $l=1, 2, \dots C$. In practice, we compute the probability of each emotion for each labeled gait in a dataset as the fraction of annotators who labeled the gait with the corresponding emotion. To perform classification, we need to convert each element in $L\samplei$ to an assignment in $\braces{0, 1}$, resulting in the multi-hot emotion label $y\samplei \in \braces{0, 1}^C$. Taking into account the subjectivity in perceiving emotions, we set an element $l$ in $y\samplei$ to 1 if $p_l > \frac{1}{C}$, \textit{i.e.}, the $\raisedth{l}$ perceived emotion has more than a random chance of being reported, and 0 otherwise. Since our classification problem is multi-class (typically, $C > 2$) as well as multi-label (as we use multi-hot labels), we use the weighted multi-class cross-entropy loss
\begin{equation}
    \mathcal{C}_{CL}\parens{y\samplei, \hat{y}\samplei} := -\sum_{l=1}^C w_l \parens{y_l}\samplei\log\parens{\hat{y}_l}\samplei
    \label{eq:cl_loss}
\end{equation}
for our classifier loss, where $\parens{y_l}\samplei$ and $\parens{\hat{y}_l}\samplei$ denote the $\raisedth{l}$ components of $y\samplei$ and $\hat{y}\samplei$, respectively. We also add per-class weights $w_l = e^{-p_l}$ to make the training more sensitive to mistakes on the rarer samples in the labeled dataset.

\subsection{Using Affective Features and Constructing Autoencoder Loss}\label{subsec:ae_loss}
Our autoencoder loss consists of three constraints: affective loss, quaternion loss, and angle loss.

\noindent\textbf{Affective loss.} Prior studies in psychology report that a person's perceived emotions can be represented by a set of scale-independent gait-based affective features~\cite{crenn2016body}. We consider the poses underlying the gaits to be made up of $J = 21$ joints (Fig.~\ref{fig:pose}). Inspired by~\cite{tanmay_emotions}, we categorize the affective features as follows:
\begin{itemize}[label=\textbullet]
    \item \textit{Angles} subtended by two joints at a third joint. For example, between the head and the neck (used to compute head tilt), the neck, and the shoulders (to compute slouching), root and thighs (to compute stride lengths), etc.
    \item\textit{Distance ratios} between two pairs of joints. For example, the ratio between the distance from the hand to the neck, and that from the hand to the root (to compute arm swings).
    \item\textit{Area ratios} formed by two triplets of joints. For example, the ratio of the area formed between the elbows and the neck and the area formed between the elbows and the root (to compute slouching and arm swings). Area ratios can be viewed as amalgamations of the angle- and the distance ratio-based features used to supplement observations from these features.
\end{itemize}
We present the full list of the $\mathcal{A} = 18$ affective features we use in Table~\ref{tab:aff_features}. We denote the set of affective features across all time steps for the $\raisedth{i}$ gait with $a\samplei \in \mathbb{R}^{\mathcal{A}\times T}$. We then constrain a subset of the embeddings learned by our encoder to map to these affective features. Specifically, we construct our embedding space to be $\mathbb{R}^{\mathcal{E}\times T}$ such that $\mathcal{E} \geq \mathcal{A}$. We then constrain the first $\mathcal{A}\times T$ dimensions of the embedding, denoted with $\hat{a}\samplei$ for the $\raisedth{i}$ gait, to match the corresponding affective features $a\samplei$. This gives our affective loss constraint:
\begin{equation}
    \mathcal{L}_{\textrm{aff}}\parens{a\samplei, \hat{a}\samplei} := \norm{a\samplei - \hat{a}\samplei}^2.
    \label{eq:aff_loss}
\end{equation}
We use affective constraints rather than providing affective features as input because there is no consensus on the universal set of affective features, especially due to cross-cultural differences~\cite{ekman_non_verbal,critical_gait_features}. Thus, we allow the encoder of our autoencoder to learn an embedding space using both data-driven features and our affective features, to improve generalizability.

\noindent\textbf{Quaternion loss.} The decoder for our autoencoder returns rotations per joint per time step as quaternions $\parens{\hat{q}_j^t}\samplei$. We then constrain these quaternions to have unit norm:
\begin{equation}
    \mathcal{L}_{\textrm{quat}}\parens{\parens{\hat{q}_j^t}\samplei} := \parens{\norm{\parens{\hat{q}_j^t}\samplei} - 1}^2.
    \label{eq:quat_loss}
\end{equation}
We apply this constraint instead of normalizing the decoder output, since individual rotations tend to be small, which leads the network to converge all its estimates to the unit quaternion.

\noindent\textbf{Angle loss.} This is the reconstruction loss for the autoencoder. We obtain it by converting the input and the output quaternions to the corresponding Euler angles and computing the mean loss between them:
\begin{equation}
    \mathcal{L}_{\textrm{ang}}\parens{D\samplei, \hat{D}\samplei} := \norm{\parens{D_X, D_Y, D_Z}\samplei - \parens{\hat{D}_X, \hat{D}_Y, \hat{D}_Z}\samplei}_F^2
    \label{eq:ang_loss}
\end{equation}
where $\parens{D_X, D_Y, D_Z}\samplei \in \bracks{0, 2\pi}^{3J\times T}$ and $\parens{\hat{D}_X, \hat{D}_Y, \hat{D}_Z}\samplei \in \bracks{0, 2\pi}^{3J\times T}$ denotes the set of Euler angles for all the joints across all the time steps for input $D\samplei$ and output $\hat{D}\samplei$, respectively, and $\norm{\cdot}_F$ denotes the Frobenius norm.

Combining Eqs.~\ref{eq:aff_loss}, \ref{eq:quat_loss} and \ref{eq:ang_loss}, we write the autoencoder loss $\mathcal{C}_{AE}\parens{\cdot, \cdot}$ as
\begin{equation}
    \mathcal{C}_{AE}\parens{D\samplei, \hat{D}\samplei} := \mathcal{L}_{\textrm{ang}}\parens{D\samplei, \hat{D}\samplei} + \lambda_{\textrm{quat}}\mathcal{L}_{\textrm{quat}} + \lambda_{\textrm{aff}}\mathcal{L}_{\textrm{aff}}
    \label{eq:ae_loss}
\end{equation}
where $\lambda_{\textrm{quat}}$ and $\lambda_{\textrm{aff}}$ are the regularization weights for the quaternion loss constraint and the affective loss constraint, respectively. To keep the scales of $\mathcal{L}_{\textrm{quat}}$ and $\mathcal{L}_{\textrm{aff}}$ consistent, we also scale all the affective features to lie in $\bracks{0, 1}$.

\begin{figure}[t]
    \centering
    \includegraphics[width=\textwidth]{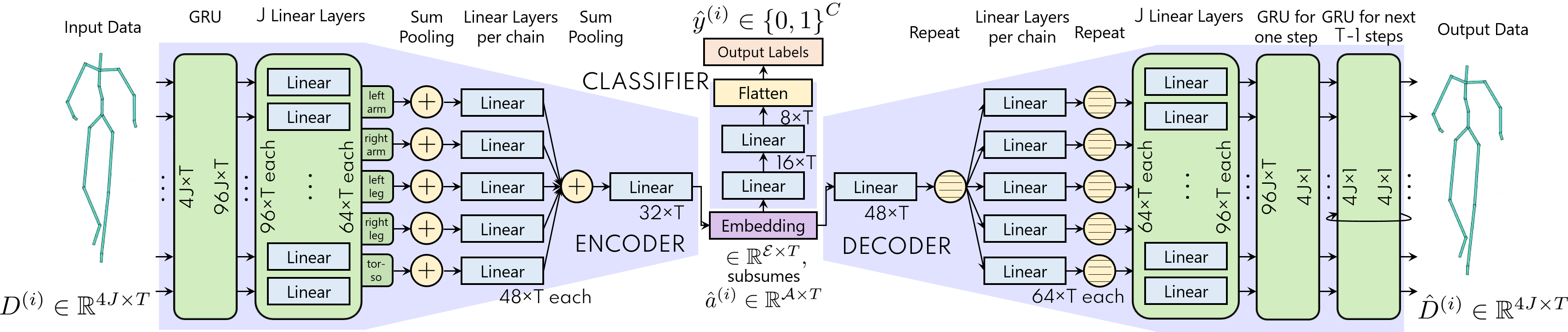}
    \caption{\smallfont\textbf{Our network for semi-supervised classification of discrete perceived emotions from gaits.} Inputs to the encoder are rotations on each joint at each time step, represented as 4D unit quaternions. The inputs are pooled bottom-up according to the kinematic chains of the human body. The embeddings at the end of the encoder are constrained to lie in the space of the mean affective features $\mathbb{R}^\mathcal{A}$. For labeled data, the embeddings are passed through the classifier to predict output labels. The linear layers in the decoder take in the embeddings and reconstruct the motion on each joint at a single time-step at the output of the first GRU. The second GRU in the decoder takes in the reconstructed joint motions at a single time step and predicts the joint motions for the next time step for $T-1$ steps.}
    \label{fig:network}
\end{figure}

\section{Network Architecture and Implementation}\label{sec:impl}
Our network for semi-supervised classification of discrete perceived emotions from gaits, shown in Fig.~\ref{fig:network}, consists of three components, the encoder, the decoder, and the classifier. We describe each of these components and then summarize the training routine for our network.

\subsection{Encoder with Hierarchical Attention Pooling}\label{subsec:encoder}
We first pass the sequences of joint rotations on all the joints through a two-layer Gated Recurrent Unit (GRU) to obtain feature representations for rotations at all joints at all time steps. We pass each of these representations through individual linear units. Following the kinematic chain of the human joints~\cite{human_kinematics}, we pool the linear unit outputs for the two arms, the two legs, and the torso in five separate linear layers. Thus, each of these five linear layers learns to focus attention on a different part of the human body. We then pool the outputs from these five linear layers into another linear layer, which, by construction, focuses attention on the motions of the entire body. For pooling, we perform vector addition as a way of composing the features at the different hierarchies.

Our encoder learns the hierarchy of the joint rotations in a bottom-up manner. We map the output of the last linear layer in the hierarchy to a feature representation in the embedding space of the encoder through another linear layer. In our case, the embedding space lies in $\mathbb{R}^{\mathcal{E}\times T}$ with $\mathcal{E} = 32$, which subsumes the space of affective features $\mathbb{R}^{\mathcal{A}\times T}$ with $\mathcal{A} = 18$, as discussed in Sec.~\ref{subsec:ae_loss}.

\subsection{Decoder with Hierarchical Attention Un-pooling}\label{subsec:decoder}
The decoder takes in the embedding from the encoder, repeats it five times for un-pooling, and passes the repeated features through five linear layers. The outputs of these linear layers are features representing the reconstructions on the five parts, torso, two arms, and two legs. We repeat each of these features for un-pooling, and then collectively feed them into a GRU, which reconstructs the rotation on every joint at a single step. A subsequent GRU takes in the reconstructed joint rotations at a single time step and successively predicts the joint rotations for the next $T-1$ time steps.

\subsection{Classifier for Labeled Data}\label{subsec:classifier}
Our classifier takes in the embeddings and passes it through a series of three linear layers, flattening the features between the second and the third linear layers. The output of the final linear layer, called ``Output Labels'' in Fig.~\ref{fig:network}, provides the label probabilities. To make predictions, we set the output for a class to be $1$ if the label probability for that class was more than $\frac{1}{C}$, similar to the routine for constructing input labels discussed in Sec.~\ref{subsec:cf_loss}.

\subsection{Training Routine}\label{subsec:train_routine}
We train using the Adam optimizer~\cite{adam} with a learning rate of $0.001$, which we decay by a factor of $0.999$ per epoch. We apply the ELU activation~\cite{elu} on all the linear layers except the output label layer, apply batch normalization~\cite{batchnorm} after every layer to reduce internal covariance-shift, and apply a dropout of $0.1$ to prevent overfitting. On the second GRU in the decoder, which predicts joint rotations for $T$ successive time steps, we use a curriculum schedule~\cite{curriculum_schedule}. We start with a teacher forcing ratio of $1$ on this GRU and at every epoch $E$, we decay the teacher forcing ratio by $\beta = 0.995$, \textit{i.e.}, we either provide this GRU the input joint rotations with probability $\beta^E$, or the GRU's past predicted joint rotations with probability $1 - \beta^E$. Curriculum scheduling helps the GRU to gently transition from a teacher-guided prediction routine to a self-guided prediction routine, thereby expediting the training process.

We train our network for $500$ epochs, which takes around $4$ hours on an Nvidia GeForce GTX 1080Ti GPU with $12$ GB memory. We use $80\%$ of the available labeled data and all the unlabeled data for training our network, and validate its classification performance on a separate $10\%$ of the labeled data. We keep the remaining $10\%$ as the held-out test data. We also observed satisfactory performance when the weights $\lambda_\textrm{quat}$ and $\lambda_\textrm{aff}$ (in Eqn.~\ref{eq:ae_loss}) lie between $0.5$ and $2.5$. For our reported performances in Sec.~\ref{subsec:experiments}, we used a value of $2$ for both.

\section{Results}\label{sec:results}
We perform experiments with the Emotion-Gait benchmark dataset~\cite{step}. It consists of 3D pose sequences of gaits collected from a variety of sources and partially labeled with perceived emotions. We provide a brief description of the dataset in Sec.~\ref{subsec:dataset}. We list the methods we compare with in Sec.~\ref{subsec:methods}. We then summarize the results of the experiments we performed with this dataset on all these methods in Sec.~\ref{subsec:experiments}, and describe how to interpret the results in Sec.~\ref{subsec:interpretation}.

\begin{table}[t]
    \begin{minipage}[t]{0.48\linewidth}
        \caption{\smallfont\textbf{Average Precision scores.} Average precision (AP) per class and the mean average precision (mAP) over all the classes achieved by all the methods on the Emotion Gait dataset. Classes are Happy (H), Sad (S), Angry (A) and Neutral (N). Higher values are better. Bold indicates best, blue indicates second best.}
        \label{tab:precision}
        \centering
        \resizebox{0.9\columnwidth}{!}{%
            \begin{tabular}{lccccc}
                \toprule
                Method & \multicolumn{4}{c}{AP} & mAP \\
                \cline{2-5}
                & H & S & A & N & \\
                \midrule
                STGCN~\cite{stgcn} \Tstrut \Bstrut & 0.98 & 0.83 & 0.42 & 0.18 & 0.61 \\
                DGNN~\cite{dgnn} \Tstrut \Bstrut & 0.98 & 0.88 & 0.73 & 0.37 & 0.74 \\
                MS-G3D~\cite{dgnn} \Tstrut \Bstrut & {\color{blue}0.98} & {\color{blue}0.88} & {\color{blue}0.75} & 0.44 & 0.76 \\
                LSTM Network~\cite{tanmay_emotions} \Tstrut \Bstrut & 0.96 & 0.84 & 0.62 & 0.51 & 0.73 \\
                STEP~\cite{step} \Tstrut \Bstrut & 0.97 & 0.88 & 0.72 & {\color{blue}0.52} & {\color{blue}0.77}  \\
                \midrule
                \textbf{Our Method} \Tstrut \Bstrut & \textbf{0.98} & \textbf{0.89} & \textbf{0.81} & \textbf{0.71} & \textbf{0.84} \\
                \bottomrule
            \end{tabular}
        }
    \end{minipage}
    \hfill
    \begin{minipage}[t]{0.48\linewidth}
        \caption{\smallfont\textbf{Ablation studies.} Comparing average precisions of ablated versions of our method. HP denotes Hierarchical Pooling, AL denotes the Affective Loss constraint. AP, mAP, H, S, A, N are reused from Table~\ref{tab:precision}. Bold indicates best, blue indicates second best.}
        \label{tab:ablation}
        \centering
        \resizebox{\columnwidth}{!}{%
            \begin{tabular}{lccccc}
                \toprule
                Method & \multicolumn{4}{c}{AP} & mAP \\
                \cline{2-5}
                & H & S & A & N & \\
                \midrule
                With only labeled data, no AL or HP \Tstrut \Bstrut & 0.92 & 0.81 & 0.51 & 0.42 & 0.67 \\
                With only labeled data, HP and no AL \Tstrut \Bstrut & 0.93 & 0.81 & 0.63 & 0.49 & 0.72 \\
                With only labeled data, AL and no HP \Tstrut \Bstrut & 0.96 & 0.86 & 0.70 & 0.51 & 0.76 \\
                With only labeled data, AL and HP \Tstrut \Bstrut & 0.97 & 0.86 & 0.72 & 0.55 & 0.78 \\
                \midrule
                With all data, no AL or HP \Tstrut \Bstrut & 0.94 & 0.83 & 0.55 & 0.48 & 0.70 \\
                With all data, HP and no AL \Tstrut \Bstrut & 0.96 & 0.85 & 0.70 & 0.60 & 0.78 \\
                With all data, AL and no HP \Tstrut \Bstrut & {\color{blue}0.97} & {\color{blue}0.87} & {\color{blue}0.76} & {\color{blue}0.65} & {\color{blue}0.81} \\
                \textbf{With all data, AL and HP} \Tstrut \Bstrut & \textbf{0.98} & \textbf{0.89} & \textbf{0.81} & \textbf{0.71} & \textbf{0.84} \\
                \bottomrule
            \end{tabular}
        }
    \end{minipage}
\end{table}

\subsection{Dataset}\label{subsec:dataset}
The Emotion-Gait dataset~\cite{step} consists of gaits collected from various sources of 3D pose sequence datasets, including BML~\cite{bml}, Human3.6M~\cite{human3.6m}, ICT~\cite{ict}, CMU-MoCap~\cite{cmu_mocap} and ELMD~\cite{elmd}. To maintain a uniform set of joints for the pose models collected from diverse sources, we converted all the models in Emotion-Gait to the $21$ joint pose model used in ELMD~\cite{elmd}. We clipped or zero-padded all input gaits to have $240$ time steps, and downsampled it to contain every $\raisedth{5}$ frame. We passed the resultant $48$ time steps to our network, we have \textit{i.e.}, $T = 48$. In total, the dataset has $3,924$ gaits of which $1,835$ have emotion labels provided by 10 annotators, and the remaining $2,089$ are not annotated. Around $58\%$ of the labeled data have happy labels, $32\%$ have sad labels, $23\%$ have angry labels, and only $14\%$ have neutral labels (more details on the project webpage).

\begin{figure}[t]
    \begin{minipage}[t]{0.56\linewidth}
        \centering
        \includegraphics[width=\columnwidth]{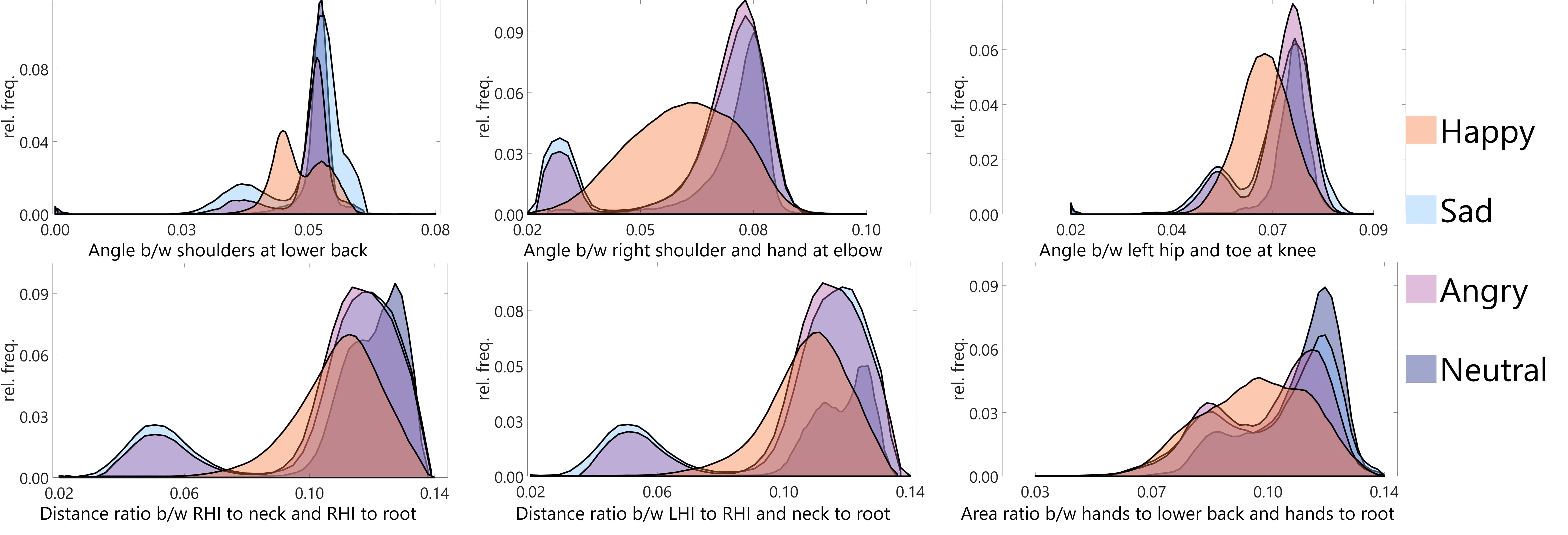}
        \caption{\smallfont\textbf{Conditional distribution of mean affective features.} Distributions of $6$ of the $18$ affective features, for the Emotion-Gait dataset, conditioned on the given classes Happy, Sad, Angry, and Neutral. Mean is taken across the number of time steps. We observe that the different classes have different distributions of peaks, indicating that these features are useful for distinguishing between perceived emotions.}
        \label{fig:aff_features}
    \end{minipage}
    \hfill
    \begin{minipage}[t]{0.4\linewidth}
        \centering
        \includegraphics[width=0.8\columnwidth]{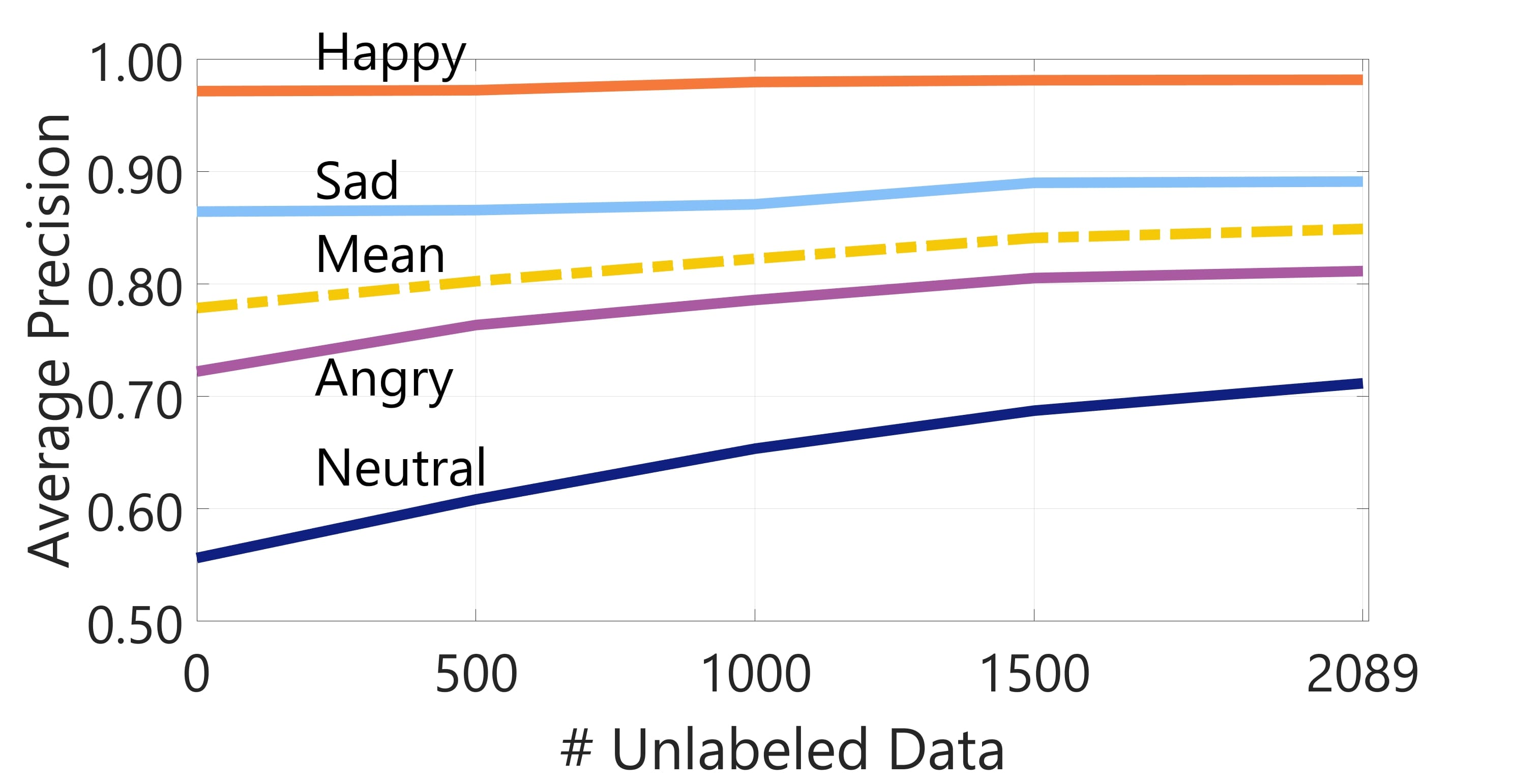}
        \caption{\smallfont\textbf{AP increases with adding unlabeled data.} AP achieved on each class, as well as the mean AP over the classes, increases linearly as we add more unlabeled data to train our network. The increment is most significant for the neutral class, which has the fewest labels in the dataset.}
        \label{fig:data_map_increase}
    \end{minipage}
\end{figure}

\noindent\textbf{Histograms of Affective Features.} We show histograms of the mean values of $6$ of the $18$ affective features we use in Fig.~\ref{fig:aff_features}. The means are taken across the $T = 48$ time steps in the input gaits and differently colored for inputs belonging to the different emotion classes as per the annotations. We count the inputs belonging to multiple classes once for every class they belong to. For different affective features, different sets of classes have a high overlap of values while values of the other classes are well-separated. For example, there is a significant overlap in the values of the distance ratio between right-hand index to the neck and right-hand index to the root (Fig.~\ref{fig:aff_features}, bottom left) for gaits belonging to sad and angry classes, while the values of happy and neutral are distinct from these. Again, for gaits in happy and angry classes, there is a high overlap in the ratio of the area between hands to lower back and hands to root (Fig.~\ref{fig:aff_features}, bottom right), while the corresponding values for gaits in neutral and sad classes are distinct from these. The affective features also support observations in psychology corresponding to perceiving emotions from gaits. For example, slouching is generally considered to be an indicator of sadness~\cite{gait_psych3}. Correspondingly, we can observe that the values of the angle between the shoulders at the lower back (Fig.~\ref{fig:aff_features}, top left) are lowest for sad gaits, indicating slouching.

\subsection{Comparison Methods}\label{subsec:methods}
We compare our method with the following state-of-the-art methods for both emotion recognition and action recognition from gaits. We choose to compare with action recognition methods because similar to these methods, we aim to learn a mapping from gaits to a set of labels (emotions instead of actions).
\begin{itemize}[label=\textbullet]
    \item \textbf{Emotion Recognition.} We compare with the network of~\cite{tanmay_emotions}, which combines affective features from gaits with features learned from an LSTM-based network taking pose sequences of gaits as input, to form hybrid feature vectors for classification. We also compare with STEP~\cite{step}, which trains a spatial-temporal graph convolution-based network with gait inputs and affective features obtained from the gaits, and then fine-tunes the network with data generated from a graph convolution-based variational autoencoder.
    \item \textbf{Action Recognition.} We compare with recent state-of-the-art methods based on the spatial-temporal graph convolution network (STGCN)~\cite{stgcn}, the directed graph neural network (DGNN)~\cite{dgnn}, and the multi-scale graph convolutions with temporal skip connections (MS-G3D)~\cite{msg3d}. STGCN computes spatial neighborhoods as per the bone structure of the 3D poses and temporal neighborhoods according to the instances of the same joints across time steps and performs convolutions based on these neighborhoods. DGNN computes directed acyclic graphs of the bone structure based on kinematic dependencies and trains a convolutional network with these graphs. MS-G3D performs multi-scale graph convolutions on the spatial dimensions and adds skip connections on the temporal dimension to model long-range dependencies for various actions.
\end{itemize}
For a fair comparison, we retrained all these networks from scratch with the labeled portion of the Emotion-Gait dataset, following their respective reported training parameters, and the same data split of $8:1:1$ as our network.

\subsubsection{Evaluation Metric}
Since we deal with a multi-class, multi-label classification, we report the average precision (AP) achieved per class, which is the mean of the precision values across all values of recall between $0$ and $1$. We also report the mean AP, which is the mean of the APs achieved in all the classes.

\begin{figure*}[t]
    \centering
    \includegraphics[width=0.9\columnwidth]{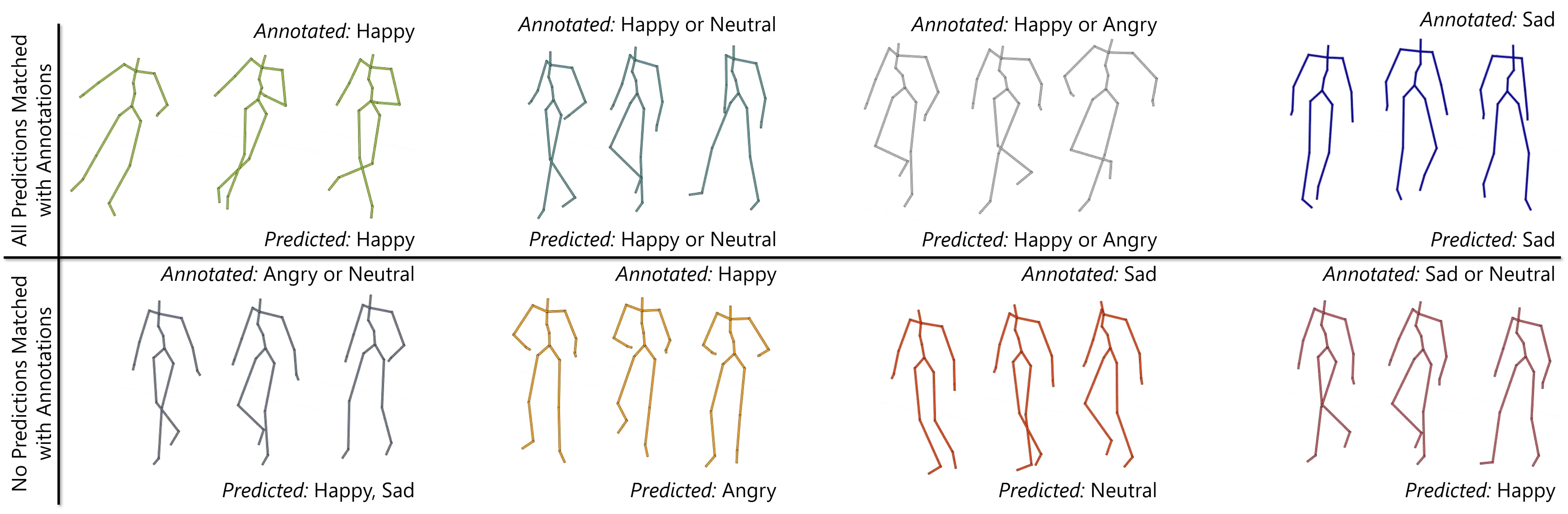}
    \caption{\smallfont\textbf{Comparing predictions with annotations.} The top row shows $4$ gaits from the Emotion-Gait dataset where the predicted labels of our network exactly matched the annotated input labels. The bottom row shows $4$ gaits where the predicted labels did not match any of the input labels. Each gait is represented by $3$ poses in temporal sequence from left to right. We observe that most of the disagreements are between either happy and angry or between sad and neutral, which is consistent with general observations in psychology.}
    \label{fig:pos_neg_examples}
\end{figure*}

\subsection{Experiments}\label{subsec:experiments}
In our experiments, we ensured that the held-out test data were from sources different from the train and validation data in the Emotion-Gait dataset. We summarize the AP and the mean AP scores of all the methods in Table~\ref{tab:precision}. Our method outperforms the next best method, STEP~\cite{step}, by around $7\%$ and outperforms the lowest-performing method, STGCN~\cite{stgcn}, by $23\%$, both on the absolute. We summarize additional results, including the interpretation of the data labels and our results in the VAD dimensions~\cite{vad}, on our project webpage.

Both the LSTM-based network and STEP consider per-frame affective features and inter-frame features such as velocities and rotations as inputs but do not explicitly model the dependencies between these two kinds of features. Our network, on the other hand, learns to embed a part of the features learned from joint rotations in the space of affective features. These embedded features, in turn, help our network predict the output emotion labels with more precision.

The action recognition methods STGCN, DGNN, and MS-G3D focus more on the movements of the leaf nodes, \textit{i.e.}, hand indices, toes, and head. These nodes are useful for distinguishing between actions such as running and jumping but do not contain sufficient information to distinguish between perceived emotions.

Moreover, given the long-tail nature of the distribution of labels in the Emotion-Gait dataset (Sec.~\ref{subsec:dataset}), all the methods we compare with have more than $0.95$ AP in the happy and more than $0.80$ AP in the sad classes, but perform much poorer on the angry and the neutral classes. Our method, by contrast, learns to map the joint motions to the affective features, which helps it achieve around $10$--$50\%$ better AP on the absolute on the angry and the neutral class while maintaining similarly high AP in the happy and the sad classes.

\subsubsection{Ablation Studies}
We also perform ablation studies on our method to highlight the benefit of each of our three key components: using hierarchical pooling (HP) (Sec.~\ref{subsec:encoder}), using the affective loss constraint (AL) (Eqn.~\ref{eq:aff_loss}), and using both labeled and unlabeled data in a semi-supervised manner (Eqn.~\ref{eq:semisup_loss}). We summarize the observations of our ablation studies in Table~\ref{tab:ablation}.

First, we train our network only on the labeled dataset by removing the decoder part of our network and dropping the autoencoder loss from Eqn.~\ref{eq:semisup_loss}. Without using either AL or HP, the network achieves an AP of $0.51$ on angry and $0.42$ on neutral, the two least populous classes. We call this our baseline network. Adding only the AL increases these two APs more from the baseline than adding only the HP. This is reasonable since hierarchical pooling helps the network learn generic differences in the pose sequences of different data, while the affective loss constraint helps the network to distinguish between pose structures specific to different perceived emotions. Adding both HP and AL increases the AP from the baseline even further. From these experiments, we can confirm that using either AL or HP improves the performance from the baseline, and their collective performance is better than their individual performances.

Next, we add in the decoder and use both labeled and unlabeled data for training our network, using the loss in Eqn.~\ref{eq:semisup_loss}. Without both AL and HP, the network now achieves an AP of $0.55$ on angry and $0.48$ on neutral, showing appreciable improvements from the baseline. Also, as earlier, adding in only the AL shows more benefit on the network's performance than adding in only the HP. Specifically, adding in only the HP produces $1\%$ absolute improvement in mean AP over STEP~\cite{step} (row 4 in Table~\ref{tab:precision}) and $17\%$ absolute improvement in mean AP over STGCN~\cite{stgcn} (row 1 in Table~\ref{tab:precision}). Adding in only the AL produces $4\%$ absolute improvement in mean AP over STEP~\cite{step} (row 4 in Table~\ref{tab:precision}) and $20\%$ absolute improvement in mean AP over STGCN~\cite{stgcn} (row 1 in Table~\ref{tab:precision}). Adding in both, we get the final version of our network, which improves on the mean AP of STEP~\cite{step} by $7\%$, and the mean AP of STGCN~\cite{stgcn} by $23\%$.

\subsubsection{Performance Trend with Increasing Unlabeled Data}
In practice, it is relatively easy to collect unlabeled gaits from videos or using motion capture. We track the performance improvement of our network as we keep adding unlabeled data to our network, and summarize the results in Fig.~\ref{fig:data_map_increase}. We observe that the mean AP improves linearly as we add more data. The trend does not indicate a saturation in AP for the angry and the neutral classes even after adding all the $2,089$ unlabeled data. This suggests that the performance of our approach can increase further with more unlabeled data. 

\subsection{Interpretation of the Network Predictions}\label{subsec:interpretation}
We show the qualitative results of our network in Fig.~\ref{fig:pos_neg_examples}. The top row shows cases where the predicted labels for a gait exactly matched all the corresponding annotated labels. We observe that the gaits with happy and angry labels in the annotation have more animated joint movements compared to the gaits with sad and neutral labels, which our network was able to successfully learn from the affective features. This is in line with established studies in psychology~\cite{vad}, which show that both happy and angry emotions lie high on the arousal scale, whereas neutral and sad are lower on the arousal scale. The bottom row shows cases where the predicted labels for a gait did not match any of the annotated labels. We notice that most disagreements arise either between sad and neutral labels or between happy and angry labels. This again follows the observation that both happy and angry gaits, higher on the arousal scale, often have more exaggerated joint movements, while both sad and neutral gaits, lower on the arousal scale, often have more reserved joint movements. There are also disagreements between happy and neutral labels for some gaits, where the joint movements in the happy gaits are not as exaggerated.

We also make an important distinction between the multi-hot input labels provided by human annotators and the multi-hot predictions of our network. The input labels capture the subjectivity in human perception, where different human observers perceive different emotions from the same gait based on their own biases and prior experiences~\cite{critical_gait_features}. The network, on the other hand, indicates that the emotion perceived from a particular gait data best fits one of the labels it predicts for that data. For example, in the third result from left on the top row in Fig.~\ref{fig:pos_neg_examples}, five of the ten annotators perceived the gait to be happy, three perceived it to be angry, and the remaining two perceived it to be neutral. Following our annotations procedure in Sec.~\ref{subsec:classifier}, we annotated this gait as an instance of both happy and angry. Given this gait, our network predicts a multi-hot label with 1's for happy and angry and 0's for neutral and sad. This indicates that the network successfully focused on the arousal in this gait, and found the emotion perceived from it to best match either happy or angry, and not match neutral and sad. We present more such results on our project webpage.

\section{Limitations and Future Work}\label{sec:limitations}
Our work has some limitations. First, we consider only discrete emotions of people and do not explicitly map these to the underlying continuous emotion space given by the VAD model~\cite{vad}. Even though discrete emotions are presumably easier to work with for non-expert end-users, we plan to extend our method to work in the continuous space of emotions, \textit{i.e.}, given a gait, our network regresses it to a point in the VAD space that indicates the perceived emotions.

Second, our network only looks at gait-based features to predict perceived emotions. In the future, we plan to combine these features with cues from other modalities such as facial expressions and body gestures, that are often expressed in tandem with gaits, to develop more robust emotion perception methods. We also plan to look at higher-level information, such as the presence of other people in the vicinity, background context, etc. that are known to influence a person's emotions~\cite{context1,context2} to further sophisticate the performance of our network.

\clearpage
%
%
\bibliographystyle{splncs04}
\bibliography{1012}

\end{document}